 \pdfoutput=1 
 
 \documentclass[pmlr,twocolumn,10pt]{jmlr} 

 \usepackage{longtable}

\usepackage{booktabs}

\usepackage{float}  
\usepackage{multirow}

\usepackage{siunitx}

\usepackage[switch]{lineno}

\usepackage{hyperref} 

\jmlrvolume{LEAVE UNSET}
\jmlryear{2024}
\jmlrsubmitted{LEAVE UNSET}
\jmlrpublished{LEAVE UNSET}
\jmlrworkshop{Machine Learning for Health (ML4H) 2024} 

\title[Long-term T2DM complication prediction]{Exploring Long-Term Prediction of Type 2 Diabetes Microvascular Complications}

 \author{
 \Name{Elizabeth Remfry} \Email{e.a.remfry@qmul.ac.uk}\\
 \Name{Rafael Henkin} \Email{r.henkin@qmul.ac.uk}\\
 \Name{Michael R Barnes} \Email{m.r.barnes@qmul.ac.uk}\\
 \addr Queen Mary University of London, UK 
 \AND
 \Name{Aakanksha Naik} \Email{aakankshan@allenai.org}\\
 \addr Allen Institute for AI, USA
}

\begin{document}
\maketitle
\begin{abstract}
Electronic healthcare records (EHR) contain a huge wealth of data that can support the prediction of clinical outcomes. EHR data is often stored and analysed using clinical codes (ICD10, SNOMED), however these can differ across registries and healthcare providers. Integrating data across systems involves mapping between different clinical ontologies requiring domain expertise, and at times resulting in data loss. To overcome this, code-agnostic models have been proposed. We assess the effectiveness of a code-agnostic representation approach on the task of long-term microvascular complication prediction for individuals living with Type 2 Diabetes. Our method encodes individual EHRs as text using fine-tuned, pretrained clinical language models. Leveraging large-scale EHR data from the UK, we employ a multi-label approach to simultaneously predict the risk of microvascular complications across 1-, 5-, and 10-year windows. We demonstrate that a code-agnostic approach outperforms a code-based model and illustrate that performance is better with longer prediction windows but is biased to the first occurring complication. Overall, we highlight that context length is vitally important for model performance. This study highlights the possibility of including data from across different clinical ontologies and is a starting point for generalisable clinical models.
\end{abstract}

\begin{keywords}
Clinical language models, electronic healthcare records, multi-label classification, disease prediction, pretrained models, type 2 diabetes, time series 
\end{keywords}

\paragraph*{Data and Code Availability}\label{para:data_code_availability} 
This study uses the Clinical Practice Research Datalink (CPRD), real-world anonymised patient data from primary care across the UK and linked to other health related registries. CPRD AURUM includes routinely collected data on 19 million patients including demographics, diagnoses, symptoms, prescriptions, referrals, lifestyle factors and tests \citep{wolf_data_2019}. Data access is subject to approval from an Independent Scientific Advisory Committee (ISAC). Code is available \texttt{\href{https://github.com/LizRem/diabetes-complications}{github.com/LizRem/diabetes-complications}}

\paragraph*{Institutional Review Board (IRB)}
The application was reviewed by an (ISAC) and the data were used under license for the current study. 

\section{Introduction}
\label{sec:intro}

Type 2 Diabetes (T2DM) is a long-term cardiometabolic condition associated with increased risk of microvascular complications; diabetic retinopathy, nephropathy and neuropathy. These complications can result in severe outcomes, such as vision loss, end stage renal disease and amputations, respectively \citep{brownrigg_microvascular_2016,khanam_microvascular_2017}. Approximately one-third of individuals living with T2DM develop at least one of these complications \citep{arnold_incidence_2022}, which in turn increases the risk of developing others \citep{deshpande_epidemiology_2008}. As various risk factors for microvascular complications are modifiable, timely identification of individuals at high risk of developing these diseases can help to inform treatment pathways and healthcare interventions \citep{khalil_diabetes_2017,lu_vascular_2023}.

Recent research has demonstrated the utility of deep learning models for disease prediction tasks due to their ability to handle messy electronic healthcare record (EHR) data which is temporal, sparse and high-dimensional \citep{hassaine_untangling_2020,wornow_shaky_2023}. Deep learning approaches for such tasks typically represent diseases as clinical codes, which requires mapping between heterogeneous clinical ontologies and manual curation or reduction of codes. Moreover, code-based representations also make such approaches less likely to generalize to unseen diseases and complications as well as across different healthcare settings. 

To address these caveats with code-based representations, our study explores a code-agnostic design taking inspiration from \citet{munoz-farre_sehr-ce_2022,hur_unifying_2022}. This approach leverages existing clinical knowledge embedded in pre-trained language models and integrates a wider range of data from across different health registries. We combine this with a multi-label approach which enables us to construct shared representations of T2DM complications, which is beneficial as complications are closely related and often share various risk factors. We explore disease prediction over short-, mid- and long-term time windows.  

\section{Related Work}
\label{sec:relatedwork}
There are a plethora of pre-trained clinical language models, however, due to data privacy very few are publicly available and those that are, come with limitations due to the heterogeneity of code ontology used in the training data \citep{wornow_shaky_2023}. 

To navigate this challenge of detaching models from the specific ontologies, research has started to utilise the natural language descriptions of the clinical codes. \citet{munoz-farre_sehr-ce_2022} utilised textual descriptors fed into an encoder only model pre-trained on clinical literature and then fine-tuned to predict various diseases. They reported improved performance compared to a model trained using traditional code embeddings. \citet{hur_unifying_2022} compared various model set ups; trained from scratch, continual pre-training and fine-tuning, on textual descriptors from MIMIC-III and eICU and found that BERT performed similarly to the models trained on clinical literature, even under different training approaches.

Our work builds on previous studies  by including a broader range of clinical data at a granular level without aggregating codes in clinical hierarchies. We include all textual descriptors within the EHR, which includes diagnoses, prescriptions, symptoms, referrals and procedures. We particularly focus on the prediction of T2DM microvascular complications over different and longer time intervals.

\section{Methods}
\label{sec:methods}

\subsection{Cohort}
We analysed EHRs from CPRD AURUM and included all individuals $\geq$ 18, permanently registered to any General Practice in London between 01/01/2010 and 01/01/2020, see Data and Code Availability  for more details.

Our dataset included 133,784 patient records, with 44,820 experiencing at least one microvascular complication \tableref{tab:characteristics}. A diagnosis of T2DM, retinopathy, neuropathy or nephropathy were identified using validated phenotype definitions and we used the first occurring diagnosis date \citep{eto_multiply-initiative_2023}. Patients with micro-vascular complications prior to a diagnosis of T2DM were excluded. 

Study entry was defined as the first EHR event until the visit prior to the first recorded complication, or the last recorded event for those without complications. We evaluated 1-, 5- and 10- years risk prediction windows post first complication. Only patients with at least 3 unique events were included.

\begin{table}[!h]
\caption{Cohort Characteristics}
\medskip
\label{tab:characteristics}
\centering
\resizebox{0.8\columnwidth}{!}{%
\begin{tabular}{@{}ll@{}}
\toprule
Characteristic                 &               \\ \midrule
Number of patients             & 133,784       \\
Total number of complications  &               \\
\quad 0                        & 88,964        \\
\quad 1                        & 33,161        \\
\quad 2                        & 9,282         \\
\quad 3                        & 2,377         \\
Number with each complication  &               \\
\quad Retinopathy              & 31,396        \\
\quad Nephropathy              & 19,595        \\
\quad Neuropathy               & 7,865         \\
Sex                            &               \\
\quad Male                     & 72,012        \\
\quad Female                   & 61,772        \\
Age at first complication (SD) & 63.06 (14.73) \\ \bottomrule
\end{tabular}%
}
\end{table}

\begin{table*}[!t]
\caption{Performance Comparison of Text- and Code-based Models}
\label{tab:performance_comparison}
    \centering
    \resizebox{0.8\textwidth}{!}{%
    \begin{tabular}{lccccc}
    \toprule
    & & \multicolumn{2}{c}{\textbf{Text-based}} & \multicolumn{2}{c}{\textbf{Code-based}} \\
    \cmidrule(r){3-4} \cmidrule(r){5-6}
     &  & \textbf{Micro-F1} & \textbf{Micro-AUPRC} & \textbf{Micro-F1} & \textbf{Micro-AUPRC} \\
    \midrule
    \textbf{1 year} &  & 0.45 \small{(0.44-0.46)} & 0.44 \small{(0.43-0.46)} & 0.43 \small{(0.42- 0.44)} & 0.40 \small{(0.38-0.41)} \\
    \textbf{5 year} &  & \textbf{0.50 (0.49-0.51)} & \textbf{0.51 (0.50-0.52)} & 0.43 (0.42-0.44) & 0.43 (0.41-0.44)\\
    \textbf{10 year} &  & 0.49 (0.48-0.50) & \textbf{0.50 (0.49-0.51)} & 0.47 (0.46-0.49) & 0.47 (0.45-0.48) \\
    \bottomrule
    \multicolumn{6}{l}{\small Note: Values in brackets represent 95\% confidence intervals, bold indicates statistical significance} \\
\end{tabular}}
\end{table*}

\vspace{-5pt}
\subsection{EHR pre-processing}
Every clinical code is associated with a textual descriptor, for example the ICD10 code \emph{E11.9} is associated with \emph{type 2 diabetes mellitus without complications}. For our text-based approach we take the textual descriptor for every event in a patient’s EHR (diagnosis, procedure, symptoms, prescription, etc.). All textual terms are then concatenated chronologically to generate text sentences for each patient. For our code-based approach, we take the clinical code and concatenate them chronologically producing a sequence of codes \appendixref{apd:first}.

\subsection{Model architecture}
We utilised a pretrained clinical language model, GatorTron-base \citep{yang_large_2022}, to encode the tokenized EHR sequences. All sequences were truncated or padded to 512 tokens, the maximum length for GatorTron. We then fine-tuned the pretrained model, one for each risk prediction window. The models consisted of a fine-tuned encoder with a single linear output layer with 3 output nodes. We split our data 80/10/10 into training, test and validation using stratified sampling to ensure the imbalance remained the same. We used weighted cross entropy due to label imbalance and report on micro F1, micro recall and micro area under the precision recall curve (AUPRC). All results are presented calculated on the held out test set. For more information on pre-processing and architecture see \appendixref{apd:first}.

We assess the variation in model performance and calculate a 95\% confidence interval (CI), by employing a bootstrap resampling technique. Using our test set of 13,314 patients we performed 1000 bootstrap iterations. Pairwise comparisons between models were conducted using a z-test approach, where the standard error of the difference was derived from the bootstrapped CIs. To account for multiple comparisons, we applied the Bonferroni correction adjusting our significance threshold.

\section{Results}
\label{sec:results}

\textbf{Code-agnostic models outperform code-based models}: Models trained on textual descriptors performed significantly better than models trained on clinical codes although not at all time windows (\tableref{tab:performance_comparison}). This suggests that there is utility in using the textual terms which may allow the model to take advantage of existing clinical knowledge.

\textbf{Models perform better over longer prediction timeframes}: the 5-year risk prediction window achieved a micro-AUPRC of 0.51 (\tableref{tab:performance_comparison}). This is likely due to the number of additional labels providing a more balanced dataset, as the longer prediction windows increases the likelihood of observing a complication.

\begin{table*}[h!]
    \centering
    \caption{F1 and Recall Scores for Microvascular Complications at 1, 5, and 10 years}
    \label{tab:mainresults}
    \resizebox{0.6\textwidth}{!}{%
    \begin{tabular}{lcccccc}
    \toprule 
            & \multicolumn{2}{c}{\textbf{Nephropathy}} & \multicolumn{2}{c}{\textbf{Retinopathy}} & \multicolumn{2}{c}{\textbf{Neuropathy}} \\ 
    \cmidrule(lr){2-3} \cmidrule(lr){4-5} \cmidrule(lr){6-7}
\textbf{Time}  & \textbf{F1}   & \textbf{Recall}   & \textbf{F1}   & \textbf{Recall}   & \textbf{F1}   & \textbf{Recall}   \\ 
    \midrule
\textbf{1 year}      & 0.39 & 0.44     & 0.51 & 0.54     & 0.22 & 0.18     \\
\textbf{5 year}      & 0.42 & 0.45     & 0.55 & 0.57     & 0.29 & 0.29     \\
\textbf{10 year}     & 0.44 & 0.51     & 0.55 & 0.55     & 0.30 & 0.30     \\ 
    \bottomrule
    \end{tabular}}
\end{table*}

\textbf{The multi-label design is biased towards first-occurring T2DM complication}:  Across all time frames, retinopathy is the highest achieving class (\tableref{tab:mainresults}), this is likely due to being the most commonly occurring first condition (in 60.19\% of cases) and the largest class. Nephropathy is the first complication in only 30.15\% of cases and neuropathy 9.67\%. As the model is only exposed to data up until the visit prior to the first complication and complications can occur at different timepoints across the life course this early data may not contain sufficient information for the model to make an accurate prediction about subsequent complications.

\textbf{Restrictions on context length affects performance}: we explored the average number of tokens in each individual's EHR (median: 2272), which falls substantially over the capability of GatorTron at 512 maximum token length. This results in the truncation of 85.37\% of sequences leading to data loss, see \appendixref{apd:first} for further exploration. In order to mimic clinicians behaviour, where they typically look at recent events first in an EHR, we mirror this by truncating from the left, removing the earliest data and preserving the most recent events \tableref{tab:truncated_left}.

\begin{table*}[ht]
    \caption{F1 and Recall Scores for Microvascular Complications and Micro-F1/AUPRC at 1, 5, and 10 years for Models Truncated Left}
    \label{tab:truncated_left}
    \centering
    \resizebox{0.8\textwidth}{!}{%
    \begin{tabular}{lcccccccc}
        \toprule
        & \multicolumn{2}{c}{\textbf{Nephropathy}} & \multicolumn{2}{c}{\textbf{Retinopathy}} & \multicolumn{2}{c}{\textbf{Neuropathy}} & \textbf{Micro-F1} & \textbf{Micro-AUPRC} \\
        \cmidrule(lr){2-3} \cmidrule(lr){4-5} \cmidrule(lr){6-7}
        \textbf{Time} & \textbf{F1} & \textbf{Recall} & \textbf{F1} & \textbf{Recall} & \textbf{F1} & \textbf{Recall} & & \\
        \midrule
        \textbf{1 year}  & 0.53 & 0.54 & 0.70 & 0.75 & 0.34 & 0.35 & 0.61 & 0.64 \\
        \textbf{5 year}  & 0.53 & 0.59 & 0.73 & 0.73 & 0.38 & 0.35 & 0.62 & 0.66 \\
        \textbf{10 year} & 0.57 & 0.61 & 0.74 & 0.75 & 0.39 & 0.40 & 0.64 & 0.69 \\
        \bottomrule
    \end{tabular}}
\end{table*}

Truncating from left led to improved performance across all prediction windows, suggesting that the EHR events recorded closer in time to a diagnosis of a complication are more important that events happening earlier. We also present the performance of pre-trained models with longer context length (4096 token length) in \appendixref{apd:first}, and indicate that shorter context lengths negatively impact performance.

\section{Discussion and future work}
\label{sec:discussion}
We present a code-agnostic method for long-term microvascular complication prediction in Type 2 Diabetes that utilises textual descriptors associated with clinical codes, unifying data across different health registries and taking advantage of pretrained language models.

\textbf{Real-world assessment of reusability of pretrained models}: Our study found that pre-trained models yielded relatively low performance on T2DM complication prediction over various time-frames despite being heralded as reusable and capable of saving time and resources. Other previous studies \citep{munoz-farre_sehr-ce_2022,hur_unifying_2022} have also yielded varying performances depending on model design, indicating that we should more thoroughly investigate reusability of pretrained models for real-world prediction tasks. Some work has tried extract more utility out of pretrained models via continual pretraining. \citet{munoz-farre_sehr-ce_2022} conducted continual pretraining using a MLM task and then fine-tuned a pretrained model which demonstrated better performance with an average AUPRC 0.61 across 4 diseases. In the future, we plan to investigate continual pretraining with span-based MLM that masks out multiple tokens representing a medical concept or phrase \citep{joshi_spanbert_2020} as in our setting multiple tokens may represent a single concept (e.g. Type 2 Diabetes).

\textbf{Incorporating data beyond text}: In our approach we limited ourselves to text descriptions, however there are additional sources of data such as numerical test results that could improve performance. For instance, \citet{hur_unifying_2022} combined both textual descriptors and numerical embeddings in a fine-tuned BERT model and achieved a 0.59 AUPRC on a multi-label disease classification task. We leave investigation of this to future work.

\textbf{Addressing context length limitations}: As we include more records from across different registries, creating a richer picture of a patient, this limits model performance as models are unable to handle longer sequences and capture long term dependencies. We plan to assess models with longer context lengths \citep{beltagy_longformer_2020}, as well as hierarchical models to further improve this \citep{li_hi-behrt_2023}. 

\textbf{Intrinsic task  difficulty}: Finally, some diseases may be clinically harder to predict using language models. \citet{li_behrt_2020} with a model trained on codes from scratch achieved AUPRC of 0.53 in a multi-label classification task across 301 diseases (classes). The performance varied, from 0.07 AUPRC for hearing loss, to 0.65 AUPRC for epilepsy although both diseases had roughly the same occurrence ratio  of 0.02. 

We believe that this work will prompt discussion around generalisable multi-purpose language models not tied to one specific healthcare setting or ontology and promote research comparing performance across different datasets. 

\acks{This work uses data provided by patients and collected by the NHS as part of their care and support. ER is funded by the Wellcome Trust Health Data in Practice (HDiP) Programme (218584/Z/19/Z). RH is funded by the AI for Multiple Long Term Conditions (AIM) Programme (NIHR203982). RM is supported by Barts Charity (MGU0504). This research utilised Queen Mary's Apocrita HPC facility, supported by QMUL Research-IT.}

\bibliography{references}

\appendix

\section{}\label{apd:first}
\subsection{Pre-processing}

Patients were only included if they were eligible for data linkage to Hospital Episode Statistics (HES) and Office for National Statistics (ONS) registries. This ensured that only patients with primary and secondary care were included. Data was pre-processed to remove duplicate events (identical rows), impossible events (dates of events that occur before birth or after deregistration), events with missing dates, or missing clinical code (events without a textual descriptor). Due to data quality issues only records between 1985 and 2020 were included \citep{wolf_data_2019}.

\begin{figure*}[htbp]
\floatconts
  {fig:input}
  {\caption{Input Format for Text and Code-based Approaches}}
  {\includegraphics[width=\textwidth]{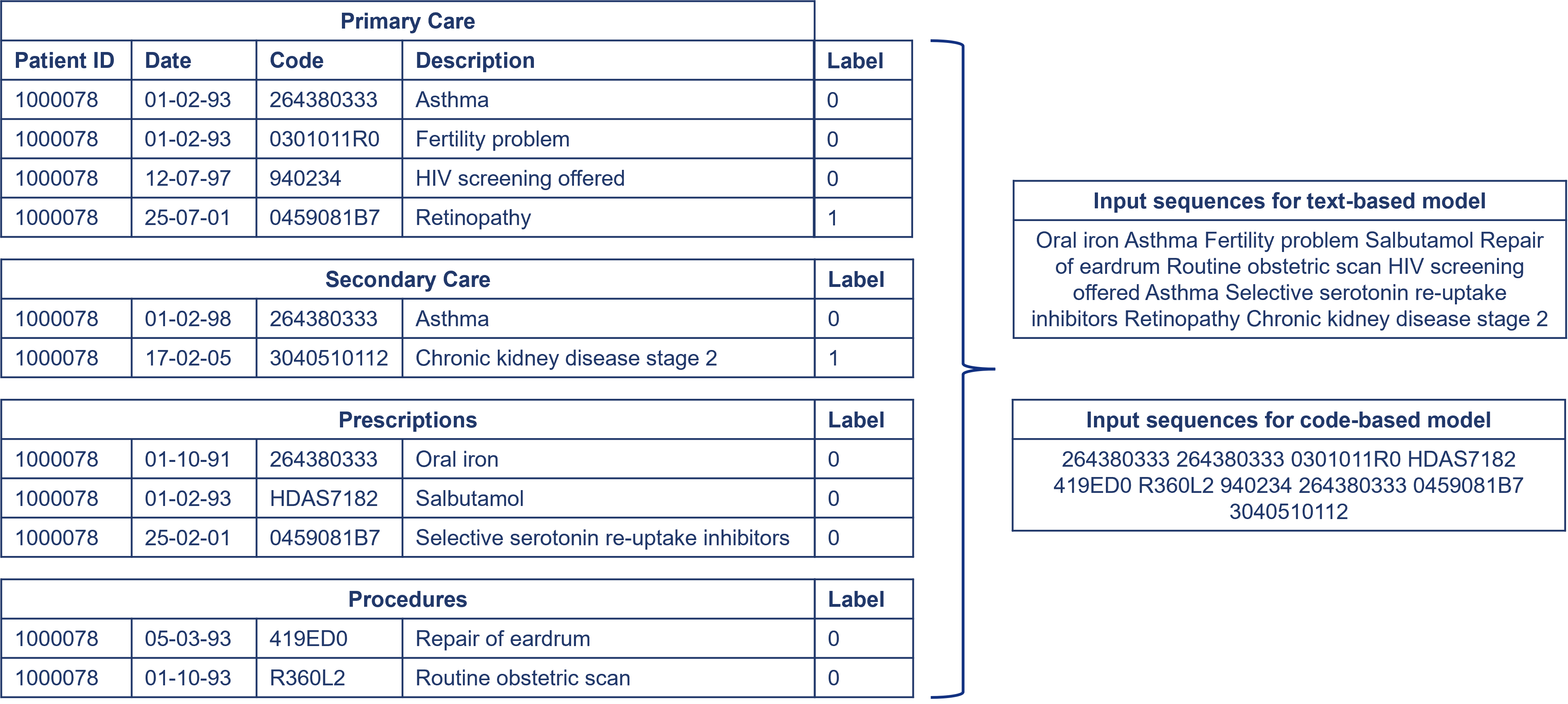}}
\end{figure*}

For the code-based model we kept the clinical codes from each of the registries, whilst for the code-agnostic models we kept the textual descriptions (\figureref{fig:input}). For data in primary care including diagnoses, symptoms, demographics etc, this follows Systematized Nomenclature of Medicine Clinical Terms (SNOMED CT), prescriptions within primary care follow the British National Formulary (BNF), within secondary care, diagnoses utilise the International Classification of Diseases, Tenth Revision (ICD10) and procedures use the OPCS Classification of Interventions and Procedures (OPCS 4). 

\subsection{Model architecture}

Gatortron-base is a smaller version of the original with 345M parameters. It was trained on scratch on 82B words of de-identified clinical notes, 6.1B words from PubMed, 2.5B words from WikiText and 0.5B words of de-identified clinical notes from MIMIC-III.

For all models, input was first tokenized and special token [CLS] added. The tokenized sequences, special tokens and positional embeddings were fed into the pretrained encoder-only model. The final hidden state of the [CLS] token was used as input to the fully connected layer. A sigmoid activation function was applied to logits to produce independent probabilities for each label (\figureref{fig:model}).

\begin{figure}[htbp]
\floatconts
  {fig:model}
  {\caption{Multi-label Approach and Model Architecture}}
  {\includegraphics[width=0.7\linewidth]{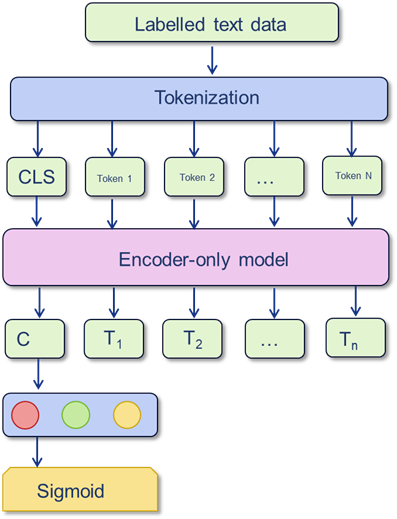}}
\end{figure}

For each model we searched for a learning rate that gave the lowest F1 score (1e-3, 2e-5, 3e-5, 4e-5, 5e-5) and fine-tuned on the entire dataset for 48000 steps with early stopping. Losses were monitored for overfitting. Models were fine-tuned on an NVidia A100 GPU. This research utilised Queen Mary's Apocrita HPC facility, supported by QMUL Research-IT \citep{king_apocrita_2017}.

\subsection{Comparison to other pretrained models}
To assess the potential benefits gained from the existing knowledge encoded in the pre-trained clinical model, GatorTron-base, we also compare to an out of domain pre-trained model, BERT-base \citep{devlin_bert_2019} trained on Wikipedia and Google Books and additionally to Biomedical-longformer-base \citep{beltagy_longformer_2020}, a model trained on abstracts from PubMed and PubMed Central articles. The Biomedical-longformer is based on the Longformer architecture, which uses an attention mechanism that scales linearly enabling a max token length of 4096. These sequences are truncated from right, as the default. 

We can see from \tableref{tab:performance_comparison_detailed} that all models perform better on text-based approaches, compared to code-based approaches. BERT performs similarly but marginally worse than GatorTron, indicating that large pre-trained models, even when not trained directly on clinical data still contain valuable knowledge. 

Biomedical-longformer was significantly better when applying a text-based approach, over a code-based approach and outperformed all other models across the prediction tasks. This suggesting that there is additional information to be gained from capturing long term dependencies in the data. However, the improvement seen on Biomedical-longformer is at the expense of additional resources and time, taking on average $\sim$20 hours compared to GatorTron at $\sim$3 hours. 

\begin{table*}[h]
\caption{Performance Comparison of Text- and Code-based Models across Different Pre-traind Models and Prediction Windows}
\label{tab:performance_comparison_detailed}
    \centering
    \resizebox{1.0\textwidth}{!}{%
    \begin{tabular}{p{0.2\textwidth}lcccc}
    \toprule
    & & \multicolumn{2}{c}{\textbf{Text-based}} & \multicolumn{2}{c}{\textbf{Code-based}} \\
    \cmidrule(r){3-4} \cmidrule(r){5-6}
     &  & \textbf{Micro-F1} & \textbf{Micro-AUPRC} & \textbf{Micro-F1} & \textbf{Micro-AUPRC} \\
    \midrule
    \textbf{BERT-base} & \textbf{1 year} & 0.44 \small{(0.43, 0.45)} & 0.42 \small{(0.40, 0.43)} & 0.42 \small{(0.41, 0.43)} & 0.39 \small{(0.37, 0.40)} \\
     & \textbf{5 year} & \textbf{0.47 \small{(0.46, 0.48)}} & \textbf{0.46 \small{(0.44, 0.47)}} & 0.43 \small{(0.43, 0.45)} & 0.42 \small{(0.41, 0.43)} \\
     & \textbf{10 year} & \textbf{0.48 \small{(0.47, 0.49)}} & \textbf{0.49 \small{(0.47, 0.50)}} & 0.46 \small{(0.45, 0.47)} & 0.46 \small{(0.44, 0.47)} \\
    \midrule
    \textbf{Biomedical-} & \textbf{1 year} & \textbf{0.56 \small{(0.55, 0.57)}} & \textbf{0.57 \small{(0.56, 0.59)}} & 0.53 \small{(0.52, 0.54)} & 0.54 \small{(0.52, 0.55)} \\
    \textbf{longformer-base} & \textbf{5 year} & \textbf{0.60 \small{(0.60, 0.61)}} & \textbf{0.63 \small{(0.62, 0.65)}} & 0.53 \small{(0.52, 0.54)} & 0.55 \small{(0.54, 0.57)} \\
     & \textbf{10 year} & \textbf{0.60 \small{(0.59, 0.61)}} & \textbf{0.64 \small{(0.63, 0.66)}} & 0.56 \small{(0.55, 0.57)} & 0.60 \small{(0.58, 0.61)} \\
    \midrule
    \textbf{Gatortron-base} & \textbf{1 year} & 0.45 \small{(0.44, 0.46)} & 0.44 \small{(0.43, 0.46)} & 0.43 \small{(0.42, 0.44)} & 0.40 \small{(0.39, 0.41)} \\
     & \textbf{5 year} & \textbf{0.50 \small{(0.49, 0.51)}} & \textbf{0.51 \small{(0.50, 0.52)}} & 0.43 \small{(0.42, 0.44)} & 0.43 \small{(0.41, 0.44)} \\
     & \textbf{10 year} & 0.49 \small{(0.48, 0.50)} & \textbf{0.50 \small{(0.49, 0.51)}} & 0.47 \small{(0.46, 0.49)} & 0.47 \small{(0.45, 0.48)} \\
    \bottomrule
    \multicolumn{6}{l}{\small Note: Values in parentheses represent 95\% confidence intervals, bold indicates statistical significance} \\
\end{tabular}}
\end{table*}

\subsection{Context length}
The median number of tokens in each individual's EHR is 2272 tokens, greater than both the maximum length of GatorTron at 512 tokens. We can see from \figureref{fig:seqlength} that many EHRs are still truncated when using Biomedical-longformer as they fall over 4096 tokens.

\begin{figure}[h]
\floatconts
  {fig:seqlength}
  {\caption{Distribution of Token Counts}}
  {\includegraphics[width=1.0\linewidth]{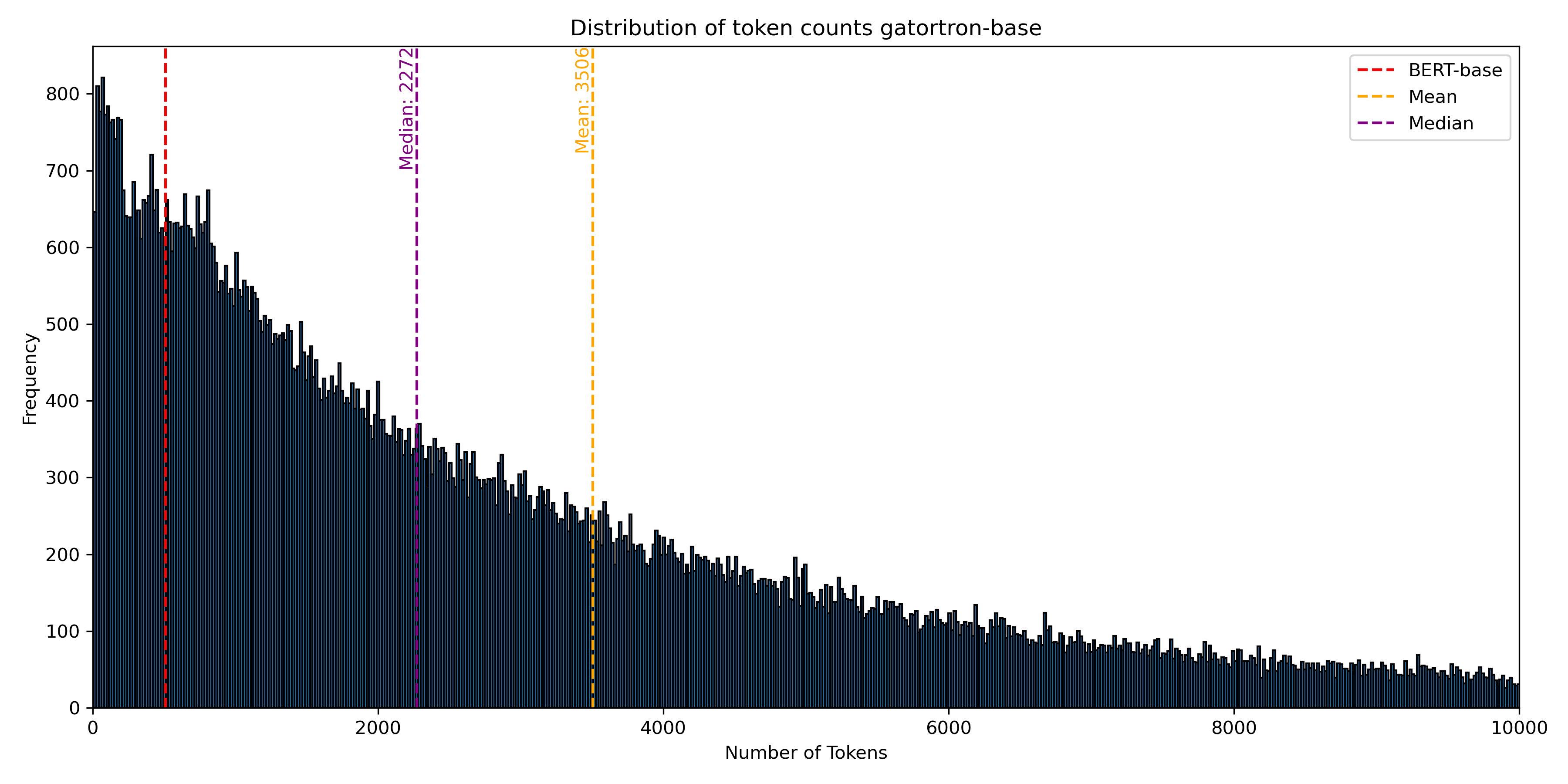}}
\end{figure}

\end{document}